\documentclass{article}
\usepackage[dblblindworkshop,final]{neurips_2026}
\workshoptitle{AI for Drug Discovery -- Bridging the Translation Gap}
\usepackage[utf8]{inputenc}
\usepackage[T1]{fontenc}
\usepackage{hyperref}
\usepackage{url}
\usepackage{booktabs}
\usepackage{amsmath}
\usepackage{amsfonts}
\usepackage{amssymb}
\usepackage{microtype}
\usepackage{xcolor}
\usepackage{graphicx}
\usepackage{siunitx}

\newcommand{\statuscheck}{\textcolor{green!55!black}{\(\checkmark\)}}
\newcommand{\statuspartial}{\textcolor{blue!70!black}{\(\triangle\)}}
\newcommand{\statuscross}{\textcolor{red!70!black}{\(\times\)}}

\title{PeptSol: A Translational Benchmark for Protected-Peptide Solubility in Drug Manufacturing}
\author{%
  Alexander Arjun Lobo \quad Marc Saouda \quad Urvi Awasthi \quad Leonid Zhukov \\
  BCG X AI Science Institute \\
  \texttt{\{Lobo.Alexander, Saouda.Marc, Awasthi.Urvi, Zhukov.Leonid\}@bcg.com}%
}

\begin{document}
\maketitle

\begin{abstract}
Liquid-phase peptide synthesis (LPPS) offers a scalable route to peptide
manufacturing, but route development still depends on empirical solvent
screens because each protected intermediate must be soluble for reaction and
recoverable for purification. The protecting groups, terminal modifications,
and noncanonical residues that govern this behaviour are poorly represented in
existing peptide property benchmarks, limiting the ability to develop and
rigorously evaluate molecular representation models for LPPS-relevant
prediction tasks. We introduce \textbf{PeptSol}, an open benchmark of
\num{1388} hand-curated protected-peptide solubility measurements with
chemistry-aware splits. The measurements span only \num{84} parent sequences
and \num{101} protected structures, and \num{92.5}\% come from a single
historical programme. Across seven frozen
molecular encoders, the best macro-F1 is \num{0.843} on binary solubility for
the historical subset under a peptide-held-out split; the harder three-class
ordinal task reaches \num{0.614} in that cell. No foundation model
consistently outperforms a baseline combining chemical fingerprints with
explicit molecular descriptors, and model rankings vary across prediction
tasks and data splits. The dataset,
evaluation protocol, and baselines establish a standard for measuring progress
on molecular representations for synthesis-aware peptide developability.
\end{abstract}

\section{Introduction}

Peptide therapeutics occupy an expanding space between small molecules and
biologics, with established applications in metabolic disease, oncology, and
infectious disease~\citep{lau2018therapeutic,muttenthaler2021trends,wang2022therapeutic}.
Discovery models usually optimize properties of the finished peptide:
activity, selectivity, permeability, stability, or toxicity. A viable drug,
however, must also be manufactured. Stepwise peptide synthesis creates a new
intermediate after every coupling, and poor reaction, solubility, or isolation
at any step can reduce yield and purity across the entire route
~\citep{bray2003manufacture,isidrollobet2019sustainability}.

Liquid-phase peptide synthesis (LPPS) offers a scalable alternative to
resin-bound synthesis. The growing chain remains in solution, enabling
in-process analysis and potentially reducing reagent and solvent excess
~\citep{okada_2013_lpps,sharma2022lpps}. This advantage creates a coupled
materials problem. A protected intermediate must dissolve sufficiently for
reaction, then crystallize or precipitate selectively for isolation
~\citep{kim_2026_oprd_crystallization}. Solvent choice is therefore not a bulk
property lookup; it depends jointly on peptide sequence, protecting groups,
terminal chemistry, and solvent environment.

Existing drug-discovery benchmarks do not represent this decision. Peptide
datasets emphasize biological function or finished-molecule properties
~\citep{huang2021tdc,li2023cycpeptmpdb,zhang2026pepbenchmark}. Protein language
models omit protecting groups and many noncanonical residues, while chemical
language models can encode the complete structure but are usually evaluated on
drug-like molecules rather than protected synthetic intermediates
~\citep{lin2023esm2,chithrananda2020chemberta,ross2022molformer}.
Consequently, a high score on a conventional peptide task does not establish
utility for route development.

\begin{figure}[t]
  \centering
  \includegraphics[width=\linewidth]{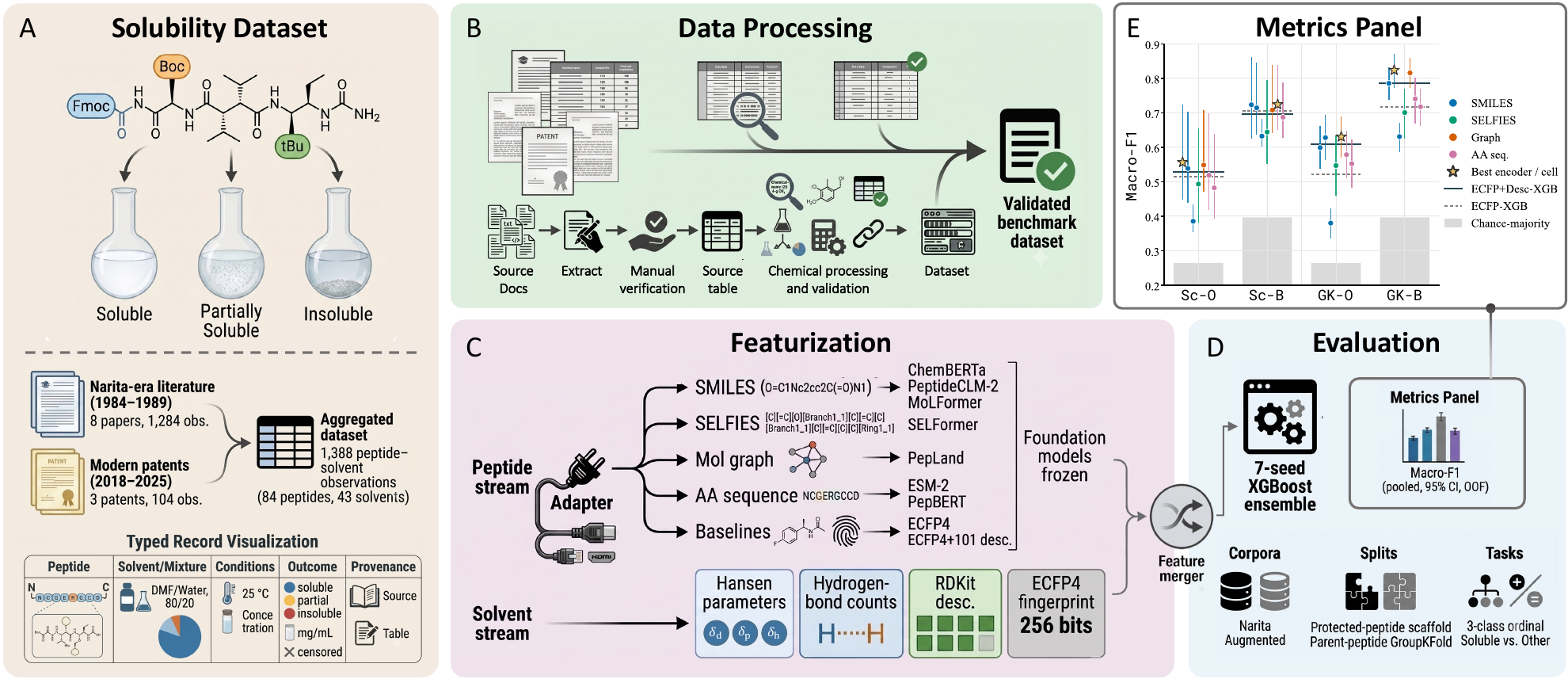}
  \caption{PeptSol links source-grounded protected-peptide--solvent records to
  fixed ordinal and binary tasks, chemistry-aware splits, and paired baseline
  comparisons.}
  \label{fig:overview}
\end{figure}

PeptSol makes this translational gap measurable. Figure~\ref{fig:overview}
summarizes the path from primary-source records to chemistry-aware evaluation.
We contribute (i) an open dataset of \num{1388} protected-peptide
solubility observations together with the deterministic curation and resolver
pipeline that builds it from source tables; (ii) an evaluation contract (fixed tasks,
chemistry-aware splits, a shared solvent block and supervised head, and
cluster-aware uncertainty) under which a claimed gain is traceable to a
change in representation; (iii) the finding that the explicit-chemistry
baseline is weakest under scaffold splitting yet second-best under peptide
grouping, so the learned-versus-explicit comparison is settled by the split; and (iv) a seven-encoder comparison against
explicit-chemistry baselines.

\section{Dataset}

\paragraph{Sources and scope.}
The corpus contains \num{1284} records transcribed from eight LPPS papers
published from 1984 to 1989
~\citep{narita_1984_prediction,narita_1984_syntheses,
narita_1984_tertiary,narita_1988_synthesis,narita_1988_hfip,
narita_1989_conformations,narita_1989_weak,narita_helix_1989} and
\num{104} records from three process patents published from 2018 to 2025
~\citep{wo2018187400_stealth,wo2024077149_lilly,wo2025243217_janssen}.
The historical series supplies a relatively consistent experimental program;
the patents add Fmoc-era intermediates and modern solvent families. The full
release covers \num{84} parent sequences, \num{101} distinct fully protected
structures, and \num{43} pure solvents or mixtures. Labels are \num{28.0}\%
insoluble, \num{6.3}\% partially soluble, and \num{65.8}\% soluble.

\paragraph{Curation and representation.}
Each observation remains linked to its primary-source table. Historical rows
were transcribed directly. A multimodal extraction pass proposed patent rows,
which were then verified one by one against the source. The version-controlled
ground-truth tables record protected sequence, solvent, conditions, verbatim
annotation, harmonized class, quantitative \si{mg/mL} value when reported,
censoring, and provenance. A deterministic build resolves peptide and solvent
names, constructs molecular structures and identifiers with RDKit
~\citep{landrum_rdkit,weininger1988smiles}, validates a typed schema, and
removes exact duplicates. Automated proposals are never treated as ground
truth without source verification. The two source regimes use different label
vocabularies (a four-level letter code in the historical tables, free-text
appearance descriptions in the patents) and each is harmonized by its own
stated convention. Appendix~\ref{sec:app-labels} publishes
both mappings verbatim, together with the curation caveats recorded on
individual rows.

\paragraph{Corpus properties.}
Measurements form a sparse peptide-by-solvent matrix rather than a randomized
screen. Qualitative labels were harmonized across protocols, the intermediate
class is small, \num{92.5}\% of records come from one historical group, and
\num{901} of \num{1388} readings are lower-bound censored; the reported
tasks consume the harmonized class rather than the censored value, as
Appendix~\ref{sec:app-censoring} details.
Section~\ref{sec:prospective} states what these properties do and do not
license.

\section{Evaluation}

\paragraph{Tasks and splits.}
The ordinal task predicts insoluble, partially soluble, or soluble, with
operational anchors at $\leq\SI{1}{mg/mL}$ and $\geq\SI{10}{mg/mL}$. The
binary task treats soluble as positive and combines the other two classes.
Macro-F1 weights the minority class equally. Two five-fold partitions measure
different generalization regimes. Peptide-GroupKFold prevents a parent peptide
from appearing in both training and test data. A Bemis--Murcko scaffold split
groups the fully protected molecular structures
~\citep{bemis_murcko_1996}. We recommend the peptide-held-out split as the
primary translational test because linear protected peptides often share a
common scaffold.

\paragraph{Models and controls.}
Seven frozen encoders span atom-level SMILES (ChemBERTa, PeptideCLM-2,
MoLFormer), SELFIES (SELFormer), molecular graphs (PepLand), and amino-acid
sequences (ESM-2, PepBERT)
~\citep{ahmad2022chemberta2,feller2026peptideclm2,
yuksel2023selformer,zhang2025pepland,guntuboina2023peptidebert}.
Every embedding is concatenated with the same \num{268}-dimensional solvent
block containing Hansen parameters, physicochemical descriptors, and an ECFP4
fingerprint. Two controls replace the embedding with a protected-peptide ECFP4
fingerprint, either alone or with \num{101} explicit sequence, protecting
group, terminal-modification, and physicochemical descriptors
~\citep{rogers2010ecfp,adamczyk2026fingerprints}.

All rows use the same seven-seed XGBoost ensemble and per-fold randomized
hyperparameter search~\citep{chen2016xgboost}. Rotational validation separates
model selection from the test fold and yields one out-of-fold prediction for
every record. We report pooled macro-F1 with \num{1000}-sample percentile
bootstrap intervals clustered by parent peptide or scaffold, matching the
split~\citep{efron1993bootstrap}. Pairwise intervals resample identical
clusters for both models.

\section{Results}

Every representation family performs above the majority-class baseline,
but foundation-model pretraining does not produce a universal winner.
Table~\ref{tab:results} reports the peptide-held-out results, the most relevant
test for a newly discovered sequence. For space, it retains the general and
peptide-specific SMILES encoders and the non-SMILES encoder with the highest
mean across these cells; the full seven-encoder grid appears in
Appendix~\ref{sec:app-results}. PeptideCLM-2
achieves the strongest binary score on the historical subset (\num{0.843}),
while PepLand leads the augmented ordinal task (\num{0.631}). Across the full
grid, paired cluster bootstrap finds no encoder significantly better than
ECFP+Desc and \num{9} significantly worse comparisons
(Table~\ref{tab:app-paired}).

\begin{table}[t]
  \centering
  \small
  \caption{Peptide-GroupKFold pooled OOF macro-F1. N-O/N-B denote Narita
  ordinal/binary; A-O/A-B denote augmented ordinal/binary.}
  \label{tab:results}
  \begin{tabular}{lrrrr}
    \toprule
    Model & N-O & N-B & A-O & A-B \\
    \midrule
    ChemBERTa   & .561 & .765 & .599 & .786 \\
    PeptideCLM-2& .614 & \textbf{.843} & .628 & \textbf{.824} \\
    PepLand     & .586 & .791 & \textbf{.631} & .816 \\
    ECFP-XGB    & .521 & .700 & .521 & .718 \\
    ECFP+Desc   & \textbf{.627} & .812 & .609 & .785 \\
    \bottomrule
  \end{tabular}
\end{table}

The task definition materially changes the apparent progress. Binary
prediction is easier because it does not require isolating the small,
censored partially-soluble class. The split also changes rankings: scaffold
partitioning can place closely related linear peptides on opposite sides of a
fold, whereas peptide grouping withholds the full sequence family. These
interactions argue against reporting one favorable aggregate score as evidence
of translational readiness.

\paragraph{Split sensitivity is systematic.}
The binary results in Table~\ref{tab:split-binary} show why the choice of split
cannot be relegated to implementation detail. PeptideCLM-2 remains the leading
encoder under peptide grouping, whereas scaffold splitting favors different
models and compresses the distinction between learned and explicit features.
The comparison does not identify one universally harder split; it identifies
two different claims. Scaffold splitting tests transfer across protected
molecular frameworks. Peptide grouping tests whether a model can support a
sequence family absent from supervised training.

\begin{table}[t]
  \centering
  \small
  \caption{Binary pooled OOF macro-F1 under scaffold (Sc) and
  peptide-GroupKFold (GK) splitting. Nar. and Aug. denote the Narita and
  augmented corpora.}
  \label{tab:split-binary}
  \begin{tabular}{lrrrr}
    \toprule
    Model & Nar.-Sc & Nar.-GK & Aug.-Sc & Aug.-GK \\
    \midrule
    ChemBERTa & .727 & .765 & .723 & .786 \\
    PeptideCLM-2 & \textbf{.739} & \textbf{.843} & .715 & \textbf{.824} \\
    PepLand & .726 & .791 & .708 & .816 \\
    ESM-2 & .731 & .797 & \textbf{.724} & .741 \\
    ECFP+Desc & .692 & .812 & .697 & .785 \\
    \bottomrule
  \end{tabular}
\end{table}

\section{From Retrospective Benchmark to Prospective Test}
\label{sec:prospective}

PeptSol is most useful as a gate between representation research and a
prospective design--make--test--learn study of peptide developability. A
retrospective score can establish that
the records contain transferable signal, that an encoder represents the
protected structure, and that gains exceed explicit molecular features. It
cannot establish that a model selects a better solvent intervention, because
the observed solvents were chosen by chemists rather than assigned at random.
The benchmark therefore supports model selection and experimental design, not
deployment.

A decisive prospective test must separate the sources of novelty that the
present splits leave confounded, listed as the transfer rows of
Table~\ref{tab:translation}, and must register predictions before measurement
so that failures enter the evaluation set. Replicates would then separate
model error from experimental variability.

The prospective endpoint should also reflect the actual decision. Binary
solubility is useful for screening but incomplete for route selection: an LPPS
solvent must support reaction and allow controlled recovery. A practical study
would record ordinal or quantitative solubility alongside coupling yield,
crystallization or precipitation success, purity, and recovered mass. PeptSol
provides the first component and the data contract needed to add the others.
The resulting evaluation would test whether a model improves the sequence of
manufacturing decisions rather than merely reproducing a historical label.

\begin{table}[t]
  \centering
  \scriptsize
  \caption{Evidence required to move from retrospective prediction to a
  translational claim. Symbols denote established (\statuscheck), partially
  tested (\statuspartial), and not established (\statuscross).}
  \label{tab:translation}
  \begin{tabular}{p{0.20\linewidth}cp{0.28\linewidth}p{0.30\linewidth}}
    \toprule
    Claim & Status & Present evidence & Required next test \\
    \midrule
    Labels are learnable & \statuscheck & OOF performance above baseline & Replication on new sources \\
    Transfer to new peptides & \statuspartial & Peptide-GroupKFold & Prospectively held-out series \\
    Transfer to new solvents & \statuscross & All aqueous records fail (App.~\ref{sec:app-errors}) & Solvent-family holdout \\
    Transfer across labs & \statuscross & \num{92.5}\% from one programme & External laboratory and protocol \\
    Transfer across eras & \statuscross & Splits are not temporal & Historical-to-patent holdout \\
    Value beyond known chemistry & \statuscross & ECFP and descriptor controls & Pre-registered paired comparison \\
    Better solvent decisions & \statuscross & Retrospective prediction only & Randomized solvent screen \\
    Manufacturing benefit & \statuscross & Solubility endpoint only & Yield, purity, and recovery endpoints \\
    \bottomrule
  \end{tabular}
\end{table}

\section{Discussion and Conclusion}

PeptSol connects molecular representation learning to an overlooked drug
development bottleneck: whether protected intermediates can be processed
through a scalable synthetic route. The results show that this behavior is
learnable from sparse public data, but also that current foundation models do
not consistently add information beyond well-chosen molecular features. For
model development, the benchmark makes the failure mode visible. For process
chemistry, it supplies a reproducible way to prioritize representations before
committing to new laboratory screens.

\paragraph{Representation transfer.}
The strongest models consume the fully protected molecular structure, which is
consistent with the importance of protecting groups and terminal chemistry.
That pattern is not sufficient to credit pretraining, however. ECFP+Desc
exposes many of the same local motifs and global physicochemical properties and
remains statistically competitive. The present study therefore supports a
narrow conclusion: frozen foundation-model embeddings contain transferable
signal for protected-peptide solubility, but they do not yet provide a reliable
advantage over explicit chemistry. End-to-end fine-tuning could change the
ranking, but it would answer a different question by combining representation
quality with adaptation capacity.

\paragraph{Evaluation design.}
Task and split definitions determine what a strong result means. The binary
task represents a practical soluble-versus-not decision but hides the
partially soluble class, whose behavior can still determine whether an
intermediate is usable. Peptide-GroupKFold directly tests transfer to a parent
sequence absent from supervised training; scaffold splitting supplies a
complementary molecular partition but can permit closely related linear
peptides across folds. Reporting both reveals this dependence, while treating
the peptide-held-out result as primary aligns the benchmark with candidate
series encountered during drug development.

\paragraph{Benchmark growth.}
The \num{1388} measurements span only \num{84} parent sequences and
\num{101} protected structures, and \num{92.5}\% come from one historical
programme, so the record count overstates the chemical coverage. The released
schema supports extension without changing the evaluation contract. New records should remain linked to source evidence, preserve raw
labels and censoring, and pass the same protected-sequence and solvent
resolvers before entering the benchmark. The highest-value additions are not
more measurements drawn from the same historical protocol; they are modern
Fmoc-era series, explicit failed experiments, replicate measurements, and
linked process outcomes.

\paragraph{What a useful model would change.}
The immediate use case is not autonomous solvent selection. It is the earlier
and narrower decision of which peptide--solvent combinations deserve scarce
experimental capacity. A model that is well calibrated on peptide-held-out
data could rank a deliberately broad solvent panel, identify cases near the
decision boundary, and allocate replicates where uncertainty is consequential.
This workflow creates a testable benefit: relative to expert screening alone,
the model should recover viable conditions with fewer experiments while
retaining chemically diverse alternatives. Evaluating that benefit requires
reporting the complete candidate panel and the selection policy, not only the
measurements ultimately performed.

The same distinction matters when the target is changed. Solubility at the
reaction stage and recovery at the isolation stage can favor different
conditions, and an intervention that improves one may impair the other. A
multi-objective extension should therefore preserve separate endpoints before
estimating a route-level utility. Collapsing coupling yield, solubility,
crystallization, purity, and solvent burden into one retrospective label would
hide the trade-offs that process chemists must resolve. PeptSol instead fixes a
single, interpretable endpoint around which these linked outcomes can be added.

\paragraph{Reproducibility as part of translation.}
The benchmark is designed so that a claimed gain can be traced to an explicit
change in representation or learning procedure. Every model receives the same
solvent block and supervised head; test rows remain outside hyperparameter
selection; predictions are pooled out of fold; and uncertainty is clustered by
the unit withheld by the split. These constraints are especially important in
the present low-data regime, where a favorable seed, fold, or treatment of
related peptides can otherwise dominate a small aggregate difference.
Releasing source locators and deterministic chemistry normalization also makes
label and structure errors auditable rather than embedding them silently in a
processed feature table.

Future comparisons should retain this contract while adding fine-tuning
controls, solvent-family and temporal holdouts, and prospective calibration and
decision-efficiency endpoints. These tests would connect representation quality
to reliable experimental information per synthesis and assay.

PeptSol does not yet justify deployment: it is retrospective, dominated by one
historical program, and lacks randomized solvents, replicates, and linked
process outcomes. What it supplies is a measurable standard: a
source-grounded corpus, a curation path that extends to new sources without
changing the evaluation, and a contract under which success on unseen peptides,
relative to explicit-chemistry baselines and with split-aware uncertainty, is a
precondition that can be checked rather than assumed.

\clearpage
\bibliographystyle{plainnat}
\bibliography{bibliography}

\clearpage
\appendix
\section{Dataset Curation and Coverage}

\subsection{Source Accounting}

The benchmark is built from human-verified source tables rather than automated
extraction output. Historical observations were transcribed directly from
eight LPPS papers. A multimodal system proposed rows from three patents, but a
curator checked every retained row against the source table. The verified
tables preserve the protected sequence, solvent, conditions, verbatim label,
harmonized class, quantitative value when reported, censoring status, and a
source locator.

\begin{table}[htbp]
  \centering
  \small
  \caption{Corpus composition. Src. denotes sources, Rec. records, Pep. unique
  parent sequences, and Solv. unique solvents or mixtures.}
  \label{tab:app-composition}
  \begin{tabular}{lcrrrr}
    \toprule
    Regime & Years & Src. & Rec. & Pep. & Solv. \\
    \midrule
    Historical literature & 1984--89 & 8 & 1284 & 67 & 25 \\
    Process patents & 2018--25 & 3 & 104 & 17 & 29 \\
    \midrule
    Full release & 1984--2025 & 11 & 1388 & 84 & 43 \\
    \bottomrule
  \end{tabular}
\end{table}

The curated tables contain \num{1405} candidate observations; \num{1388}
produce schema-supported protected structures. Four Lilly rows contain the
unsupported token \texttt{dnp}, and thirteen Janssen rows contain constructs
such as \texttt{7Me}, \texttt{2-Boc-ea}, or \texttt{Pen*}. These records fail
protected-sequence parsing and are excluded. No historical observation is
excluded. The deterministic importer makes unsupported chemistry explicit
rather than silently mapping it to an approximate structure.

\subsection{Class and Solvent Coverage}

The patent records broaden solvent and protecting-group chemistry without
materially changing the marginal class balance. Tables~\ref{tab:app-class} and
\ref{tab:app-solvents} provide the counts used to define benchmark scope.

\begin{table}[htbp]
  \centering
  \small
  \caption{Class counts in the two benchmark views.}
  \label{tab:app-class}
  \begin{tabular}{lrrr}
    \toprule
    Corpus & Insoluble & Partially soluble & Soluble \\
    \midrule
    Historical (\num{1284}) & 366 & 82 & 836 \\
    Augmented (\num{1388}) & 388 & 87 & 913 \\
    \bottomrule
  \end{tabular}
\end{table}

\begin{table}[htbp]
  \centering
  \small
  \caption{Record counts by solvent family.}
  \label{tab:app-solvents}
  \begin{tabular}{lrr@{\qquad}lrr}
    \toprule
    Family & Hist. & Aug. & Family & Hist. & Aug. \\
    \midrule
    Polar aprotic & 454 & 480 & Ether & 72 & 91 \\
    Other & 264 & 265 & Aromatic & 50 & 55 \\
    Protic & 208 & 237 & Aqueous mixture & 0 & 8 \\
    Chlorinated & 159 & 160 & Aqueous & 0 & 4 \\
    Ester & 77 & 88 & & & \\
    \bottomrule
  \end{tabular}
\end{table}

\subsection{Label Harmonization}
\label{sec:app-labels}

Every curated row stores the verbatim annotation printed in its source table
next to the harmonized class, so the mapping is auditable rather than implicit.
The two regimes use different vocabularies and are harmonized by separate,
explicitly registered conventions rather than by a shared heuristic.

\paragraph{Historical letter scheme.}
The Narita-era tables report a four-level letter code at a single assay
concentration of \SI{10}{mg/mL}. Table~\ref{tab:app-narita-labels} gives the
convention, which fixes the harmonized class, the assay temperature,
and whether the reading is censored. Letters A and B both denote dissolution at
the assay concentration and are therefore recorded as lower-bound readings;
they differ in the temperature at which dissolution was achieved. Letters C and
D denote incomplete dissolution and are uncensored.

\begin{table}[htbp]
  \centering
  \small
  \caption{Historical letter convention. The assay concentration is
  \SI{10}{mg/mL} for every row; $n$ counts records entering the benchmark.}
  \label{tab:app-narita-labels}
  \begin{tabular}{llrlr}
    \toprule
    Verbatim & Harmonized class & Assay \si{\celsius} & Censoring & $n$ \\
    \midrule
    \texttt{A} & Soluble           & 25 & Lower bound & 697 \\
    \texttt{B} & Soluble           & 80 & Lower bound & 139 \\
    \texttt{C} & Partially soluble & 80 & Uncensored  &  82 \\
    \texttt{D} & Insoluble         & 80 & Uncensored  & 366 \\
    \bottomrule
  \end{tabular}
\end{table}

\paragraph{Patent free-text annotations.}
The three process patents report visual appearance rather than a graded code.
Table~\ref{tab:app-patent-labels} lists every distinct phrase that enters the
benchmark. The nine phrases are mutually consistent across assignees: no phrase
maps to different classes in different sources. Terms denoting a single
homogeneous phase map to soluble, terms denoting a persistent second phase map
to insoluble, and intermediate terms map to partially soluble.

\begin{table}[htbp]
  \centering
  \small
  \caption{Patent appearance annotations. St., Li., and Ja. denote the Stealth,
  Lilly, and Janssen applications; $n$ counts records entering the benchmark.}
  \label{tab:app-patent-labels}
  \begin{tabular}{llrl}
    \toprule
    Verbatim annotation & Harmonized class & $n$ & Sources \\
    \midrule
    \texttt{clear solution}        & Soluble           & 63 & St., Li., Ja. \\
    \texttt{soluble}               & Soluble           & 10 & St. \\
    \texttt{clear yellow solution} & Soluble           &  4 & Li. \\
    \texttt{gel}                   & Partially soluble &  3 & Li. \\
    \texttt{hazy solution}         & Partially soluble &  2 & Li. \\
    \texttt{poorly soluble}        & Insoluble         &  9 & St. \\
    \texttt{slurry}                & Insoluble         &  6 & Li., Ja. \\
    \texttt{cloudy}                & Insoluble         &  4 & Li. \\
    \texttt{suspension}            & Insoluble         &  3 & Ja. \\
    \bottomrule
  \end{tabular}
\end{table}

\paragraph{Operational anchors on rows without a quantitative value.}
The $\leq\SI{1}{mg/mL}$ and $\geq\SI{10}{mg/mL}$ anchors define the intended
quantitative meaning of the three classes, but they do not generate any label
in the present release. The ordinal task returns a record's stored class
whenever one is present and bins a quantitative value only when the class is
absent. All \num{1405} curated rows carry an explicit class, so the numeric
cutoffs are never evaluated for the released benchmark. No historical row
carries a quantitative measurement: the \SI{10}{mg/mL} figure attached to those
rows is the assay concentration, an experimental condition, not a measured
value. The anchors therefore fix the scale on which the classes are
interpreted and govern future quantitative-only records; they are not a second
labeling path for the rows released here.

\paragraph{Recorded curation caveats.}
Historical rows also store the curation judgement that produced them, so
ambiguous cells remain visible after harmonization
(Table~\ref{tab:app-label-caveats}). The largest category is the shared
solvent column: several source tables print one letter for a group of solvents
such as \texttt{DMF/DMA/NMP}, which the build expands into one record per
component solvent carrying the same label. In \num{89} rows the source defines
the letter only relative to an unstated reference temperature, so no assay
temperature is pinned. A smaller set of cells prints two letters, or is
overridden by a table footnote that distinguishes individual solvents from the
group default.

\begin{table}[htbp]
  \centering
  \small
  \caption{Curation annotations recorded on the \num{1284} historical rows.}
  \label{tab:app-label-caveats}
  \begin{tabular}{lr}
    \toprule
    Recorded annotation & Rows \\
    \midrule
    Direct transcription, no annotation      & 866 \\
    Label shared across a grouped solvent column & 270 \\
    Assay temperature not pinned in source   &  89 \\
    Footnote overrides the base label        &  28 \\
    Printed cell carries two letters         &  23 \\
    Other curation note                      &   8 \\
    \bottomrule
  \end{tabular}
\end{table}

\subsection{Censored Readings}
\label{sec:app-censoring}

A reading is censored when the source establishes a bound rather than a value.
Table~\ref{tab:app-censoring} reports how much of the release is censored and
how censoring interacts with the presence of a numeric value.

\begin{table}[htbp]
  \centering
  \small
  \caption{Censoring in the release. Quant. counts records carrying a numeric
  \si{mg/mL} value; Both counts records that are censored and carry one.}
  \label{tab:app-censoring}
  \begin{tabular}{lrrrr}
    \toprule
    Regime & Records & Censored & Quant. & Both \\
    \midrule
    Historical literature & 1284 & 836 &  0 &  0 \\
    Process patents       &  104 &  65 & 87 & 61 \\
    \midrule
    Full release          & 1388 & 901 & 87 & 61 \\
    \bottomrule
  \end{tabular}
\end{table}

\paragraph{How censored rows enter the reported tasks.}
Both reported tasks consume the harmonized class, never the numeric reading.
The ordinal task returns the stored class and the binary task collapses that
class, so a censored record is treated identically to an uncensored record
carrying the same class. Neither task reads the censoring flag, and the flag is
not supplied to the model as a feature. Censoring therefore cannot shift a
reported macro-F1. The release retains the flag so that a censored regression
can be fitted later, but no such result is reported here: only \num{87} of
\num{1388} records carry a numeric value at all and \num{61} of those are
censored, which is too small and too heavily censored to support a
quantitative benchmark.

\paragraph{Why a censored value is never treated as a point measurement.}
A censored reading constrains the true value from one side only. For the
historical rows labelled A or B, the source establishes that the intermediate
dissolved at the \SI{10}{mg/mL} assay concentration, so the true solubility is
at least \SI{10}{mg/mL} and may be far higher; the experiment does not locate
it. Recording such a row as a point value of \SI{10}{mg/mL} would assert a
measurement the source never made, place every one of those records exactly on
the $\geq\SI{10}{mg/mL}$ anchor, and understate solubility by an unknown
margin. The release avoids this structurally rather than by convention.
\num{840} of the \num{901} censored records carry no numeric value, so there is
nothing to mistake for a measurement. For the remaining \num{61}, the ordinal
and binary tasks never read the numeric field, and the one task that exposes it
returns the bound together with its censoring flag, so a consumer cannot
silently reduce a lower bound to an equality.

\section{Representation and Evaluation Details}

\subsection{Information Available to Each Encoder}

The encoder roster deliberately mixes representations with unequal access to
chemistry. SMILES, SELFIES, and molecular-graph encoders receive the fully
protected structure, including protecting groups, terminal modifications, and
supported noncanonical residues. ESM-2 and PepBERT receive only the parent
amino-acid sequence. Their substitution layer maps supported noncanonical
residues to canonical analogues or \texttt{X}; protecting and terminal groups
have no token. Sequence-model performance must therefore be interpreted as a
lower-information control rather than a matched architectural comparison.

\begin{table}[htbp]
  \centering
  \scriptsize
  \caption{Frozen-encoder configurations used in the main benchmark.}
  \label{tab:app-adapters}
  \begin{tabular}{lllrr}
    \toprule
    Encoder & Input & Pooling & Max length & Embedding \\
    \midrule
    ChemBERTa-77M & SMILES & mean & 512 & 384 \\
    PeptideCLM-2 & SMILES & mean & 512 & 512 \\
    MoLFormer-XL & SMILES & mean & 202 & 768 \\
    SELFormer & SELFIES & mean & 512 & 768 \\
    PepLand & graph & node mean & --- & 300 \\
    ESM-2 & AA sequence & mean & 1024 & 640 \\
    PepBERT & AA sequence & mean & 52 & 320 \\
    \bottomrule
  \end{tabular}
\end{table}

\subsection{Design Rows}

Each peptide--solvent observation becomes one feature row. Encoder rows use a
frozen peptide embedding; the classical rows use a \num{2048}-bit radius-two
Morgan fingerprint, optionally followed by \num{101} explicit peptide
descriptors. All rows append the same solvent block
(Table~\ref{tab:app-solvent-block}). Mixtures use a fraction-weighted average
of component features, which does not capture nonlinear solvent interactions.

\begin{table}[htbp]
  \centering
  \small
  \caption{Solvent feature block appended to every model input.}
  \label{tab:app-solvent-block}
  \begin{tabular}{lrl}
    \toprule
    Sub-block & Dimensions & Contents \\
    \midrule
    Hansen parameters & 3 & $\delta_D,\delta_P,\delta_H$ \\
    Hydrogen bonding & 2 & donors and acceptors \\
    RDKit descriptors & 7 & LogP, TPSA, mass, and related features \\
    Morgan fingerprint & 256 & radius two \\
    \midrule
    Total & 268 & \\
    \bottomrule
  \end{tabular}
\end{table}

\subsection{Splits, Selection, and Uncertainty}

Both splitters use five folds and seed \num{42}. Fold $i$ is held out for
testing, fold $(i+1)\bmod 5$ is used for validation, and the remaining folds
train the supervised head. The scaffold split groups the Bemis--Murcko
framework of the fully protected molecule; peptide-GroupKFold groups the
parent sequence. No grouping key crosses folds.

The common XGBoost head uses a seven-seed ensemble. Within each fold, eight
randomly selected hyperparameter combinations are evaluated on the rotational
validation fold (Table~\ref{tab:app-hp}). The selected configuration is then
fit across ensemble seeds. Temperature is excluded because historical
compounds were retested hot only after failing to dissolve cold; temperature
therefore follows the outcome in that protocol.

\begin{table}[htbp]
  \centering
  \small
  \caption{Per-fold randomized hyperparameter search space.}
  \label{tab:app-hp}
  \begin{tabular}{ll}
    \toprule
    Hyperparameter & Candidate values \\
    \midrule
    Maximum depth & $\{3,4,6,8\}$ \\
    Learning rate & $\{0.05,0.1,0.2\}$ \\
    Trees & $\{100,200,400\}$ \\
    Class weighting & none, balanced, inverse \\
    \bottomrule
  \end{tabular}
\end{table}

All paper metrics are computed from raw pooled out-of-fold probabilities; no
calibrator is fit or applied. Confidence intervals use \num{1000}
cluster-level percentile bootstrap resamples with seed \num{0}. The cluster
matches the held-out grouping rule. Pairwise comparisons reuse identical
sampled clusters for both models.

\section{Extended Results and Limitations}
\label{sec:app-results}

\subsection{Complete Peptide-Held-Out Results}

The compact main-text table retains the general and peptide-specific SMILES
encoders and the non-SMILES encoder with the highest mean across the four
GroupKFold cells. Table~\ref{tab:app-all-encoders} reports all seven frozen
encoders under the primary peptide-held-out evaluation. All models use the same
solvent block, supervised head, folds, and hyperparameter-selection procedure.

\begin{table}[htbp]
  \centering
  \small
  \caption{Complete Peptide-GroupKFold pooled OOF macro-F1. N-O/N-B denote
  Narita ordinal/binary; A-O/A-B denote augmented ordinal/binary.}
  \label{tab:app-all-encoders}
  \begin{tabular}{lrrrr}
    \toprule
    Model & N-O & N-B & A-O & A-B \\
    \midrule
    ChemBERTa    & .561 & .765 & .599 & .786 \\
    PeptideCLM-2 & .614 & \textbf{.843} & .628 & \textbf{.824} \\
    MoLFormer    & .574 & .753 & .597 & .730 \\
    SELFormer    & .506 & .675 & .547 & .701 \\
    PepLand      & .586 & .791 & \textbf{.631} & .816 \\
    ESM-2        & .603 & .797 & .579 & .741 \\
    PepBERT      & .558 & .748 & .552 & .718 \\
    \midrule
    ECFP-XGB     & .521 & .700 & .521 & .718 \\
    ECFP+Desc    & \textbf{.627} & .812 & .609 & .785 \\
    \bottomrule
  \end{tabular}
\end{table}

The point-estimate leaders do not establish an advantage over explicit
chemistry. Table~\ref{tab:app-paired} reports the paired comparison between the
highest-scoring encoder in each cell and ECFP+Desc. Each interval includes
zero. Across all seven encoders, two corpora, two tasks, and two splits, none
of the \num{56} paired intervals is significantly positive and \num{9} are
significantly negative.

\begin{table}[htbp]
  \centering
  \small
  \caption{Paired macro-F1 difference between the highest-scoring frozen
  encoder and ECFP+Desc in each Peptide-GroupKFold cell. Intervals are
  \num{95}\% cluster-bootstrap confidence intervals.}
  \label{tab:app-paired}
  \begin{tabular}{llllr}
    \toprule
    Corpus & Task & Encoder & $\Delta$ & \num{95}\% CI \\
    \midrule
    Narita & Ordinal & PeptideCLM-2 & $-.013$ & $[-.059,.030]$ \\
    Narita & Binary & PeptideCLM-2 & $.032$ & $[-.015,.081]$ \\
    Augmented & Ordinal & PepLand & $.021$ & $[-.043,.091]$ \\
    Augmented & Binary & PeptideCLM-2 & $.039$ & $[-.004,.081]$ \\
    \bottomrule
  \end{tabular}
\end{table}

\subsection{Majority-Class Baselines}

The majority-class baseline predicts the training-fold majority class. Its
macro-F1 is below the uniform-random ternary reference because macro-F1 assigns
zero recall to unpredicted classes (Table~\ref{tab:app-floor}). Every learned
representation clears this floor in every benchmark cell.

\begin{table}[htbp]
  \centering
  \small
  \caption{Majority-class baseline macro-F1.}
  \label{tab:app-floor}
  \begin{tabular}{lrr}
    \toprule
    Corpus & Ordinal & Binary \\
    \midrule
    Narita & .263 & .394 \\
    Augmented & .265 & .397 \\
    \bottomrule
  \end{tabular}
\end{table}

\subsection{Feature-Block Ablation}
\label{sec:app-ablation}

Each design row concatenates a peptide block and a solvent block, so a score
alone does not say which block carries the signal. Table~\ref{tab:app-ablation}
separates them. All three conditions reuse the same folds, supervised head,
hyperparameter search, seeds, and bootstrap; only the blocks reaching the head
differ. The solvent-only condition never invokes an encoder and is therefore
adapter-independent by construction.

\begin{table}[htbp]
  \centering
  \small
  \caption{Feature-block ablation, pooled OOF macro-F1 under
  peptide-GroupKFold. Solvent-only uses the \num{268}-dimensional solvent
  block with no peptide representation.}
  \label{tab:app-ablation}
  \begin{tabular}{lrrrr}
    \toprule
    Encoder & N-O & N-B & A-O & A-B \\
    \midrule
    \multicolumn{5}{l}{\emph{(a) Solvent block only: no peptide representation}} \\
    \quad (no encoder) & .421 & .620 & .380 & .631 \\
    \addlinespace
    \multicolumn{5}{l}{\emph{(b) Peptide block only: no solvent features}} \\
    \quad ChemBERTa    & .454 & .703 & .486 & .713 \\
    \quad PeptideCLM-2 & .444 & .724 & .495 & .707 \\
    \quad MoLFormer    & .425 & .672 & .463 & .642 \\
    \quad SELFormer    & .409 & .536 & .452 & .636 \\
    \quad PepLand      & .475 & .738 & .472 & .754 \\
    \quad ESM-2        & .438 & .668 & .477 & .663 \\
    \quad PepBERT      & .420 & .620 & .478 & .670 \\
    \addlinespace
    \multicolumn{5}{l}{\emph{(c) Peptide + solvent: the benchmark configuration}} \\
    \quad ChemBERTa    & .561 & .765 & .599 & .786 \\
    \quad PeptideCLM-2 & \textbf{.614} & \textbf{.843} & .628 & \textbf{.824} \\
    \quad MoLFormer    & .574 & .753 & .597 & .730 \\
    \quad SELFormer    & .506 & .675 & .547 & .701 \\
    \quad PepLand      & .586 & .791 & \textbf{.631} & .816 \\
    \quad ESM-2        & .603 & .797 & .579 & .741 \\
    \quad PepBERT      & .558 & .748 & .552 & .718 \\
    \bottomrule
  \end{tabular}
\end{table}

Solvent identity alone is a substantial predictor. With no peptide
representation the head reaches macro-F1 \num{0.620} on the Narita binary
task, against a majority-class floor of \num{0.394}
(Table~\ref{tab:app-floor}). Across the four peptide-held-out cells the
solvent block alone recovers between \num{31} and \num{55}\% of the distance
from that floor to the best encoder. Encoder gains should therefore be read
against the solvent baseline rather than against chance.

The peptide block nonetheless contributes in every cell. Averaged over the
four cells, adding the peptide block to the solvent features improves on
solvent-only by \num{0.094} (SELFormer) to \num{0.214} (PeptideCLM-2).
Neither block suffices alone: peptide-only sits below the combination for
every encoder, and below solvent-only in several cells. The two are
complementary, consistent with solubility being a joint property of the
peptide and the solvent rather than a property of either.

\subsection{Error Analysis}
\label{sec:app-errors}

This section locates the errors behind the primary peptide-held-out result.
All figures come from the saved out-of-fold predictions of a leading
encoder, PeptideCLM-2, on the augmented ordinal task under peptide-GroupKFold;
no model is retrained. Overall accuracy is \num{82.9}\%.

\begin{table}[htbp]
  \centering
  \small
  \caption{Pooled out-of-fold confusion matrix. Rows are the curated class,
  columns the predicted class.}
  \label{tab:app-confusion}
  \begin{tabular}{lrrrr}
    \toprule
    & Insoluble & Partially & Soluble & Recall \\
    \midrule
    Insoluble & 289 & 6 & 93 & .745 \\
    Partially soluble & 21 & 13 & 53 & .149 \\
    Soluble & 56 & 8 & 849 & .930 \\
    \bottomrule
  \end{tabular}
\end{table}

\paragraph{The intermediate class carries the error.}
Per-class F1 is \num{0.767} insoluble, \num{0.890} soluble, and \num{0.228}
partially soluble. Only \num{13} of \num{87} partially-soluble records are
recovered; \num{53} are called soluble and \num{21} insoluble. Because
macro-F1 weights the three classes equally, this single class accounts for
most of the gap between the ordinal and binary scores reported throughout the
paper, and it is the mechanism behind the task asymmetry noted in the
abstract. It is also the smallest and most weakly defined class
(Section~\ref{sec:app-labels}).

\paragraph{Errors are asymmetric toward soluble.}
The model over-predicts solubility in both directions: \num{93} insoluble
records are called soluble against \num{56} in the reverse direction, and the
partially-soluble class collapses toward soluble rather than insoluble. For a
screening use case this is the costlier direction, since a false soluble call
sends an intermediate into a solvent that will not dissolve it.

\begin{table}[htbp]
  \centering
  \small
  \caption{Error rate by stratum. Strata with fewer than \num{30} records are
  pooled into the trailing row of each block.}
  \label{tab:app-errors}
  \begin{tabular}{lrr}
    \toprule
    Stratum & $n$ & Error rate \\
    \midrule
    \multicolumn{3}{l}{\emph{Solvent family}} \\
    \quad Polar aprotic & 480 & .179 \\
    \quad Other         & 265 & .196 \\
    \quad Protic        & 237 & .114 \\
    \quad Chlorinated   & 160 & .137 \\
    \quad Ether         &  91 & .132 \\
    \quad Ester         &  88 & .182 \\
    \quad Aromatic      &  55 & .182 \\
    \quad Aqueous and aqueous mixtures & 12 & \textbf{1.000} \\
    \addlinespace
    \multicolumn{3}{l}{\emph{C-terminal group}} \\
    \quad OBzl  & 962 & .160 \\
    \quad OPac  & 207 & .198 \\
    \quad OH    & 111 & .171 \\
    \quad COOH  &  48 & .083 \\
    \quad NH$_2$&  32 & \textbf{.469} \\
    \quad Other &  28 & .143 \\
    \addlinespace
    \multicolumn{3}{l}{\emph{Peptide length (residues)}} \\
    \quad $\leq$7   & 630 & .171 \\
    \quad 8--13     & 516 & .194 \\
    \quad 14--19    &  89 & .135 \\
    \quad 20--31    &  30 & .233 \\
    \quad $\geq$32  & 123 & .081 \\
    \addlinespace
    \multicolumn{3}{l}{\emph{Curated class}} \\
    \quad Soluble           & 913 & .070 \\
    \quad Insoluble         & 388 & .255 \\
    \quad Partially soluble &  87 & .851 \\
    \bottomrule
  \end{tabular}
\end{table}

\paragraph{Aqueous solvents fail completely.}
Every one of the \num{12} aqueous and aqueous-mixture records is
misclassified. These enter only through the patent extension
(Table~\ref{tab:app-solvents}). This indicates poor performance on this small
subset. Because the evaluation holds out peptides rather than solvent
families, it does not establish failure on previously unseen solvents.

\paragraph{Errors vary with C-terminal chemistry and peptide length.}
Among C-terminal groups the amide is hardest at \num{0.469}, roughly three
times the benzyl-ester rate of \num{0.160}, though on only \num{32} records.
Error against peptide length is not monotone: it is highest for
\num{20}--\num{31} residues (\num{0.233}), the smallest bin with only
\num{30} records, and lowest for \num{32} or more residues (\num{0.081}).
These descriptive differences do not establish an intrinsic length effect.

\subsection{Interpretive Limits}

The released data support retrospective prediction, not causal solvent
recommendation. Most measurements originate from one historical research
group; the modern extension contains three patent assignees. Protocols and
qualitative labels differ across sources, experimental replicates are
unavailable, and quantitative values are frequently lower-bound censored. The
peptide--solvent matrix is sparse and reflects chemist choices. Finally,
solubility does not directly measure coupling yield, crystallization, purity,
or recovery. These limitations motivate the prospective design in the main
paper rather than undermining the narrower benchmark contribution.

\end{document}